\documentclass[conference]{IEEEtran}

\usepackage{amsmath,amssymb,amsfonts}

\usepackage{graphicx}          
\usepackage{booktabs}
\usepackage{threeparttable}
\usepackage{tabularx}
\usepackage{multirow}
\usepackage[caption=false,font=footnotesize]{subfig}  

\usepackage{algorithm}         
\usepackage[noend]{algpseudocode}  

\usepackage[T1]{fontenc}
\usepackage[utf8]{inputenc}
\usepackage{microtype}

\usepackage{xcolor}
\usepackage{framed}
\usepackage[framemethod=TikZ]{mdframed}
\usepackage{tikz}              
\usepackage{tcolorbox}
\usepackage{multicol}
\usepackage{float}
\usepackage{afterpage}
\usepackage{placeins}
\usepackage{adjustbox}
\usepackage{enumitem}
\usepackage{soul}
\usepackage{pifont}
\usepackage{newunicodechar}
\usepackage[caption=false,font=footnotesize]{subfig}

\newunicodechar{≥}{$\geq$}

\usepackage{cite}

\usepackage{todonotes}

\usepackage[hidelinks]{hyperref}

\def\BibTeX{{\rm B\kern-.05em{\sc i\kern-.025em b}\kern-.08em
    T\kern-.1667em\lower.7ex\hbox{E}\kern-.125emX}}
\begin{document}

\title{A Data-Driven RetinaNet Model for Small Object Detection in Aerial Images}

\author{\IEEEauthorblockN{Zhicheng Tang, Jinwen Tang, Yi Shang}
\IEEEauthorblockA{\textit{Department of Electrical Engineering and Computer Science} \\
\textit{University of Missouri} \\
Columbia, MO, USA \\
\textit{robert.tang.robot@gmail.com}; \textit{\{jt4cc, shangy\}@umsystem.edu}}
}

\maketitle

\begin{abstract}
In the realm of aerial imaging, the ability to detect small objects is pivotal for a myriad of applications, encompassing environmental surveillance, urban design, and crisis management. Leveraging RetinaNet, this work unveils DDR-Net: a data-driven, deep-learning model devised to enhance the detection of diminutive objects. DDR-Net introduces novel, data-driven techniques to autonomously ascertain optimal feature maps and anchor estimations, cultivating a tailored and proficient training process while maintaining precision. Additionally, this paper presents an innovative sampling technique to bolster model efficacy under limited data training constraints. The model's enhanced detection capabilities support critical applications including wildlife and habitat monitoring, traffic flow optimization, and public safety improvements through accurate identification of small objects like vehicles and pedestrians. DDR-Net significantly reduces the cost and time required for data collection and training, offering efficient performance even with limited data. Empirical assessments over assorted aerial avian imagery datasets demonstrate that DDR-Net markedly surpasses RetinaNet and alternative contemporary models. These innovations advance current aerial image analysis technologies and promise wide-ranging impacts across multiple sectors including agriculture, security, and archaeology.
\end{abstract}


\section{Introduction}
Aerial image analysis plays a important role in many applications such as environmental monitoring, urban planning, agricultural management, and disaster response and recovery. High-resolution aerial imagery captured by drones provides wide-ranging views, supplying rich datasets for monitoring, analysis, and informed decision-making. Yet, the large amount of data generated from aerial surveys present a significant challenge for manual analysis. Therefore, the application of Machine Learning (ML) techniques, especially state-of-the-art deep learning models, is essential to automate the data processing pipeline, extract valuable insights, and speed up decision-making processes.

A key to the success of applications of ML, especially deep learning models, in aerial image analysis is having ample annotated datasets for training robust models. However, collecting such datasets is often prohibitively time-consuming and expensive. The diversity and complexity of aerial imagery further increase the data requirement for training models capable of generalizing well to unseen data, thus slowing the progress towards fully automated aerial image analysis systems.

Several strategies have been suggested in literature to address the data scarcity issue. Among these, transfer learning and data augmentation stand out, as they utilize existing knowledge and synthetic data augmentation to enhance model training with limited data \cite{montserrat2017training, kisantal2019augmentation,zoph2020learning}. Despite these efforts, there's still a considerable room for improvement, especially in developing architectures that are  adept at learning from limited data.

In this paper, we present a new architecture based on RetinaNet\cite{lin2017focal}, named DDR-Net, to improve how our model performs when trained with a small amount of data. Unlike existing models like RetinaNet, DDR-Net takes advantages of the specific features of small objects such as birds in aerial images and can reach a good performance with a limited amount of training examples. The main contributions of the paper are as follows:
\begin{enumerate}
    \item Propose a new method to determine appropriate sizes of feature maps and reduce the number of negative samples needed to speed up model training.
    \item Propose a new method to automatically choose the number and size of anchor boxes, which are aligned well with  real bounding boxes.
    \item Propose a new  clustering method for data sampling, aiming to get balanced representative training data.
\end{enumerate}
Our extensive experiments on multiple public datasets show that DDR-Net performed better than RetinaNet and some existing methods.

The rest of this paper is organized as follows: Section II presents related work of object detection in aerial images. Section III describes the proposed DDR-Net architecture and proposed new methods. Section IV presents the experimental setup and results. Finally, Section V concludes the paper.

\section{Related work}

Object detection is a common task in analyzing aerial images, supporting various real-world applications, such as investigating forest conservation areas \cite{zhang2022deep,sivanandam2022tree,ferreira2020individual} or monitoring waterfowl \cite{tang2023new,tang2021suas,pagano2009estimating}.
A lot of these applications use one-stage object detectors, i.e.  deep learning models for end-to-end object detection such as RetinaNet, due to their fast speeds. Moreover, their simple configurable structures make the models easy to setup. One-stage detectors have been adaptable to different  use cases.

\paragraph{Optimizing Performance with Limited Training Data}
The acquisition of substantial training data, particularly high-quality annotations, is a pivotal step in developing robust object detection models. However, when constrained by the amount of available data, employing data augmentation techniques becomes widely used to enhance a deep learning model's learning capability and, consequently, its performance. The work by \cite{kaur2021data} provides a comprehensive review of existing data augmentation techniques tailored for object detection tasks. Data augmentation techniques, such as geometric transformations and color space augmentations, artificially expand the existing dataset, thereby enriching a model's learning environment and improving its performance even with limited data. Data augmentation is particularly beneficial in real-world scenarios such as medical imaging, where data is often scarce. 

\paragraph{Adapting to Varying Target Scales}
The tasks of object detection in aerial imagery often encounter a broad range of object sizes. Several methods have been proposed to enhance model performance across diverse object scales. One such method is  adaptive anchor calculation \cite{ren2020real,wang2022improved}, which is commonly employed in anchor-based methods. In this technique, the system iterates through the training dataset to determine the optimal anchor settings that align with the data. This method notably boosts the speed and accuracy of object detection by choosing appropriate anchor boxes through a process like the K-means clustering algorithm, thereby creating sets of adaptive anchor boxes that better match the object size distribution in the dataset \cite{mehtab2021flexinet}.

Moreover, Adaptive Anchor Box Optimization (AABO) has been introduced as a way to redefine the shapes and sizes of anchor boxes, veering away from pre-defined configurations, and allowing more flexibility in dealing with varied object scales. This is especially pertinent since pre-defined anchor configurations may not always match well with the object size distributions in specific datasets \cite{kaur2021data}.

\paragraph{Refining Proposals During Training}
Another approach to improve model performance is reducing the number of proposals used during loss calculation during model training. In the original setup of RetinaNet, a matching threshold is set based on  IOU (Intersection over Union) between  real bounding box and  anchor boxes to lessen the number of anchors involved in the final calculation. This threshold might change with different training data. 
\cite{xu2021adaptive} proposed a Multi-Level Feature Fusion Module (MFFM) and a Multi-Scale Feature Fusion Unit (MFFU) to tackle the problem of not using features well in pedestrian detection, an idea that can be applied to detecting objects from aerial images. \cite{yang2003feature} proposed a parallel feature fusion technique using complex vectors for a joined-up feature representation, significantly improving classification performance when compared to the usual serial feature fusion methods. This way of representing data could be very helpful in dealing with different object sizes and complex backgrounds in aerial imagery. Moreover, how data is represented when sampling training data is key for effective learning, highlighting the importance of smart data sampling strategies.

\section{Proposed Methods}
Traditional deep learning models for object detection employ preset anchors with predefined sizes and aspect ratios, aiming to adapt to a wide range of scale variance. However, in aerial images of certain targets, such birds, target sizes exhibit lesser variance, tending to fall within a specific range determined by the Ground Sampling Distance (GSD) of the image, also known as the scale of the image. Furthermore, substantial target information can be used when detecting specific object types, although the scale may vary significantly due to differences in data collection processes such as flying altitude and camera specifications of drones. We propose DDR-Net, a new architecture based on RetinaNet, to take advantage of the useful information.

\begin{figure}
    \centering
    \includegraphics[width=0.95\linewidth]{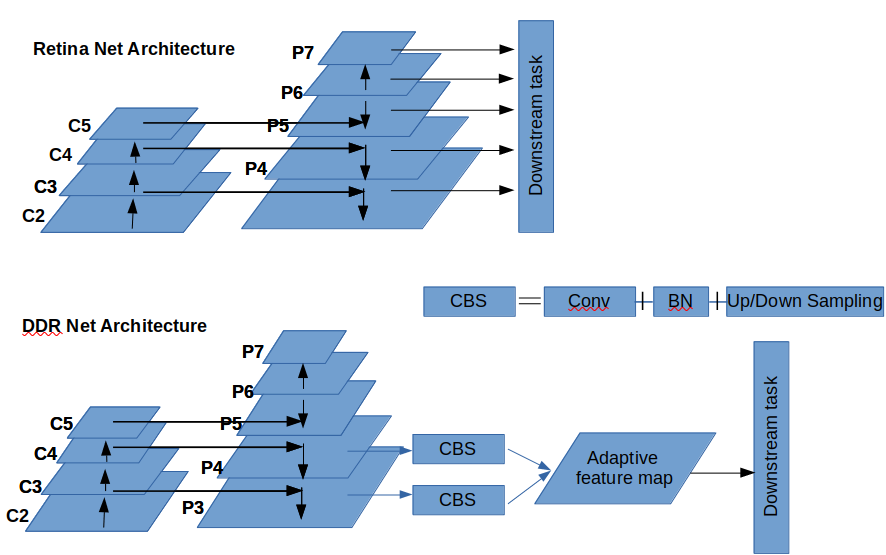}
    \caption{DDR-Net architecture.}
    \label{fig:DDR-Net architecture}
\end{figure}

The architecture of DDR-Net is designed to utilize training data in guiding architectural decisions, aiming to reduce computational overhead while enhancing model performance in both training from scratch and fine-tuning scenarios. DDR-Net has three new parts: Adaptive Feature Map Construction, Anchor Box Estimation, and ML-based Image Sampling. The architecture of DDR-Net, as compared to that of RetinaNet, is shown in Fig. \ref{fig:DDR-Net architecture}.

DDR-Net's design embraces the inherent data structures and features of aerial imagery and requires fewer training samples to achieve more accurate and efficient object detection, particularly when dealing with variations in image scale arising from differing data collection parameters. 


\subsection{Adaptive Feature Map Construction}
Conventional models like the Feature Pyramid Network (FPN) generate predictions from various feature layers. When running a pre-trained model of RetinaNet on certain targets, most predictions for objects might come from the anchors in one layer, such as the P4 layer, as shown in Fig. \ref{fig:DDR-Net architecture}. However, there are problems when fine-tuning this model on similar objects at different scales. For example, smaller objects that fit better with the P3 layer may not gain much from pre-trained information geared towards P4. This disconnect stems from the scale shift and different anchor alignments between layers.

Unlike  traditional FPN,  DDR-Net introduces an adaptive feature map estimation module to tackle scale variations. Integrated with a fusion layer, this module dynamically merges multiple feature maps, scaling them to an optimal size. This way, useful features from earlier training can be effectively transferred to a new dataset, despite  scale differences. The fusion layer, a unique aspect of DDR-Net, is designed to maximize the usefulness of pre-trained model features during fine-tuning, even with differing target sizes.

\begin{figure}
\centering
\includegraphics[width=0.95\linewidth]{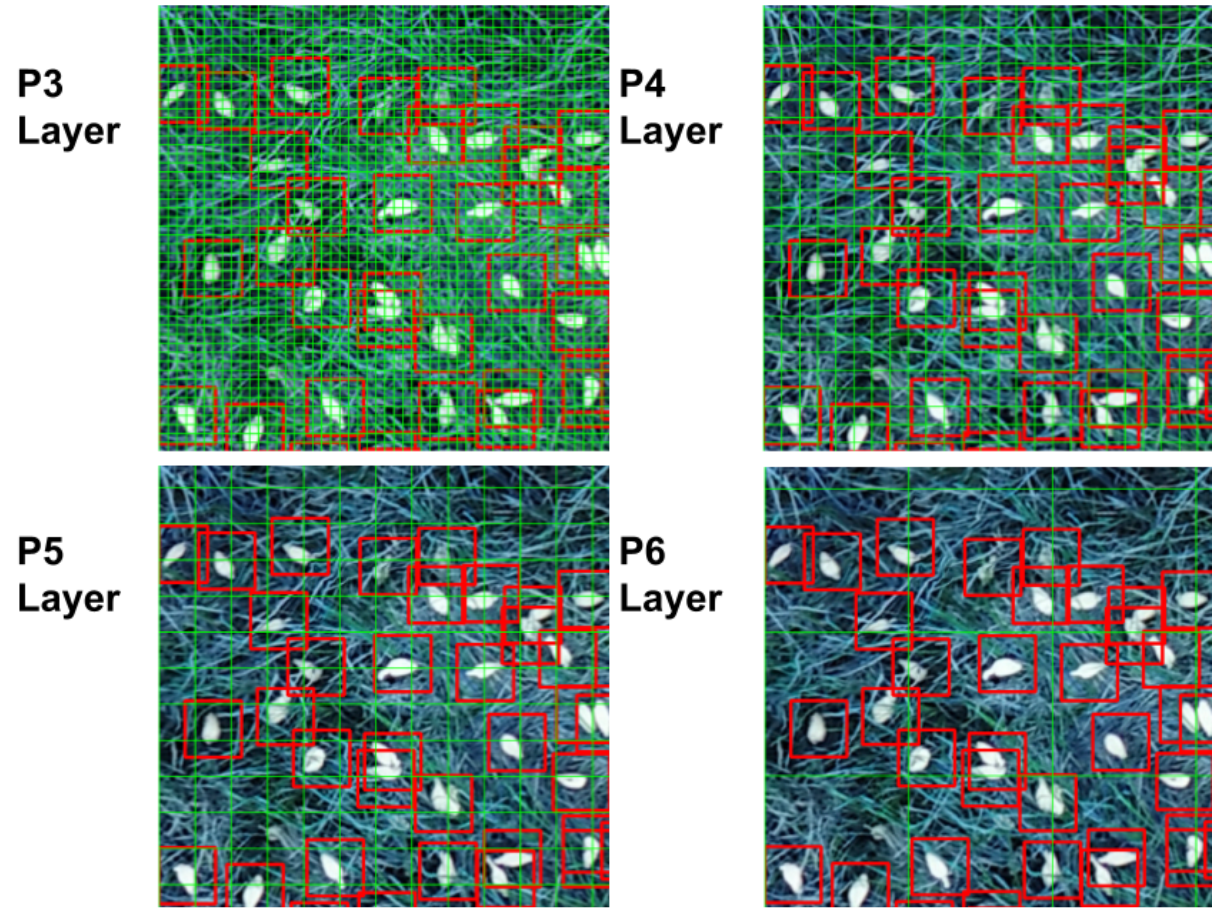}
\caption{An example of feature map sizes from P3 to P6 of RetinaNet. The green grids represent  anchor points at each feature map layer.}
\label{fig:featuremap size example}
\end{figure}

DDR-Net also addresses overfitting or underfitting issues, which are  common in aerial imagery where object sizes are more uniform due to the fixed camera position. It dynamically picks the best feature map dimensions based on the training dataset, aiming to choose sizes that provide enough anchor boxes for predictions while cutting down on negative samples. 

The procedure for finding the optimal feature map size is shown in Algorithm \ref{alg:optimal-feature-map-size}. The Matching Score Calculation method is shown in Algorithm \ref{alg:feature-map-score}. Using the Everglade dataset \cite{weinstein2022general} as an example, we calculated the feature map score with a weighting factor of -1 and penalty factor of 0.0001. The resulting scores for pyramid levels P3 to P7 of RetinaNet were: 40020.7, 44785.7, 44018.8, 38753.3, and 21990.1, respectively. Among these, P4 stood out as the best, determining the fused feature map size. Fig. \ref{fig:featuremap size example} shows an example that illustrates  feature map size from P3 to P6 of RetinaNet. Ideally, we want each anchor to match no more than a few bounding boxes (defined by the anchor estimation process) without including too many negative anchors. By observing how each annotation fits within the grid, it is evident that the P4 layer generates the best result.

This adaptive feature map technique enhances DDR-Net's adaptability and versatility across a variety of aerial image scenarios, making it useful for accurate object detection and smooth fine-tuning across diverse datasets.

\begin{algorithm*}
\caption{Identify Optimal Feature Map Size: Iterates through candidate feature map sizes, records annotation overlaps with different feature map sizes, and selects the best feature map size based on the score calculated in \textit{CalculateFeatureMapScore}.}
\label{alg:optimal-feature-map-size}
\begin{algorithmic}
\small
\Procedure{FindOptimalFeatureMapSize}{$imageAnnotations, featureMapSizes$}
    \State \textbf{Input:}
    \Statex \hspace{1cm} $imageAnnotations$: A list of annotations in the image.
    \Statex \hspace{1cm} $featureMapSizes$: A set of feature map sizes for each feature map.
    \State $featureMapData \gets \text{empty dictionary}$
    \For{each $level$ in $featureMapSizes$}
        \For{each $annotation$ in $imageAnnotations$}
            \State $overlappedGrid \gets \Call{FindOverlappedGrid}{annotation, level}$
            \State $featureMapData[level][overlappedGrid]$ $update\  by +1$
        \EndFor
    \EndFor
    \State $featureSizeScores \gets \Call{CalculateFeatureMapScore}{featureMapData, numAnchors}$
    \State $optimalFeatureMapSize \gets \text{argmax}(featureSizeScores)$
    \State \Return $optimalFeatureMapSize$
\EndProcedure
\end{algorithmic}
\end{algorithm*}

\begin{algorithm*}
\caption{Calculate Feature Map Score: Computes the score for each feature map size based on the number of overlaps on each feature map grid. Utilizes a weighting factor to measure grids with overlaps exceeding the number of anchors per grid, and a penalty factor to penalize grids with no matching annotations.}
\label{alg:feature-map-score}
\begin{algorithmic}
\small
\Procedure{CalculateFeatureMapScore}{$featureMapData, numAnchors$}
    \State \textbf{Input:}
    \Statex \hspace{1cm} $featureMapData$: A dictionary with feature map levels and overlapped annotations data.
    \Statex \hspace{1cm} $numAnchors$: Number of anchors obtained from anchor estimation process.
    \State $penalty\_factor \gets 0.0001, \, weighting\_factor \gets -1$
    \State $metrics \gets \text{empty dictionary}$
    \For{each $level, freqs$ in $featureMapData$}
        \State $desired\_overlap\_score \gets \sum_{1 \leq key \leq numAnchors} freqs[key]$
        \State $excessive\_overlap\_score \gets \sum_{key > numAnchors} freqs[key] \cdot (key - numAnchors) \cdot weighting\_factor$
        \State $penalty \gets \text{number of grids without overlapped} \cdot penalty\_factor$
        \State $metrics[level] \gets desired\_overlap\_score + excessive\_overlap\_score - penalty$
    \EndFor
    \State \Return $metrics$
\EndProcedure
\end{algorithmic}
\end{algorithm*}

\subsection{Anchor Box Estimation}
Traditional object detection models typically employ preset anchors with predefined sizes and aspect ratios to adapt to a wide range of scale variances. However, in aerial images of certain applications, such as bird detection, the variance in target sizes is often lesser due to similar ground sampling distances (GSD) of the images, also referred to as the scale of the images. For such scenarios, we propose the estimation of anchor box sizes based on  training data, replacing the preset anchor boxes.

\begin{figure}[htbp]
    \centering
    \includegraphics[width=0.9\linewidth]{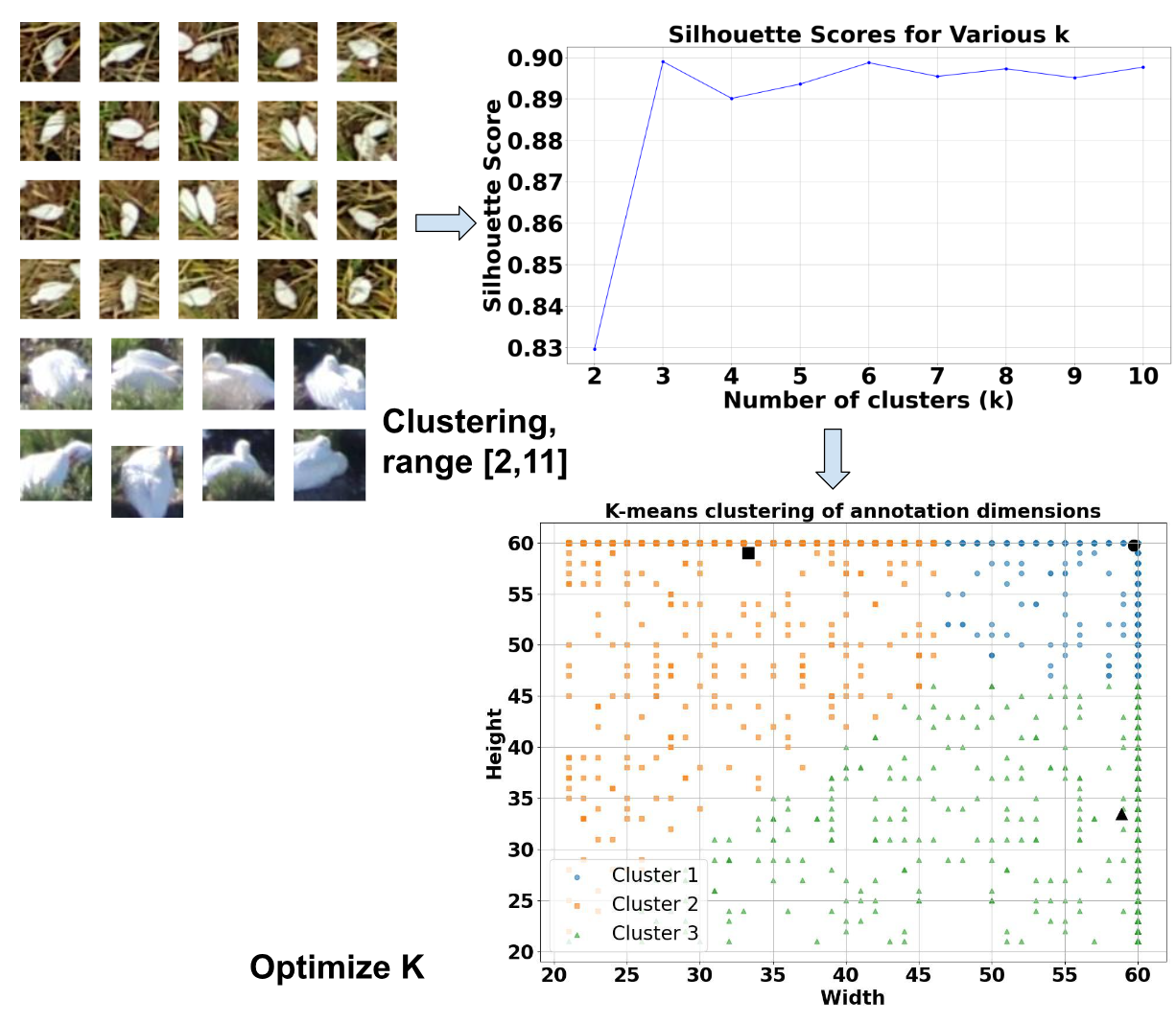}
    \caption{An example of estimating anchor sizes using k-means clustering on training data. }
    \label{fig:anchor estimate example}
\end{figure}

We use  K-means clustering algorithm to group the bounding boxes from the training dataset into clusters. The silhouette score is employed to guide the selection of optimal number of clusters. Initially, we collect all ground-truth bounding boxes and stores their widths and heights. Next, we record the silhouette score for number of clusters ranging from 2 to 6, representing the number of clusters being tested. Finally, we select the cluster number that maximizes the silhouette score. Fig. \ref{fig:anchor estimate example} shows an example where three clusters are the best.

The proposed method for estimating anchor box sizes has two  benefits. One is the reduction in effort required for calculating the bounding box regression loss. This is achieved by customizing the anchor boxes based on the distribution of ground-truth bounding boxes which in turn minimizes the regression loss between the anchor boxes and the actual bounding boxes. The second benefit is the facilitation of bridging the scale variance between different targets during transfer learning. This is further explained in the adaptive feature map section where features from different layers are merged into a single feature map.

Let's use a simple example to illustrate the benefits by using two different bird datasets: one from the Everglades dataset with a bounding box size of \(50 \times 50\), and the other from the seabirdwatch dataset with a size of \(20 \times 20\)~\cite{weinstein2022general}.

\begin{itemize}
    \item In the usual way of calculating anchor regression, the equations below are used to find the loss on width and height (assuming best fit anchor boxes),
    \begin{equation}
    \Delta w  = \log\left(\frac{w}{w_{\text{anchor}}}\right)= \log\left(\frac{50}{64}\right) \approx -0.358,
    \end{equation}
    \begin{equation}
    \Delta h  = \log\left(\frac{h}{h_{\text{anchor}}}\right) = \log\left(\frac{20}{16}\right) \approx 0.223,
    \end{equation}
    where \( w \) and \( h \) are the width and height of the real bounding box, while \( w_{\text{anchor}} \) and \( h_{\text{anchor}} \) are the width and height of the predicted anchor box.
    
    \item In our method of estimating anchor boxes, a clustering algorithm gives us anchor boxes sized \(50 \times 50\) and \(20 \times 20\) respectively. Using the same equations as before, the loss on the anchor width and height becomes zero, a perfect match.
\end{itemize}

This example shows how our way of estimating anchor boxes can lower the regression loss and handle different target sizes well during transfer learning.

Also, in a transfer learning scenario, if we want to fine-tune  models pre-trained using the Everglades dataset to  the seabirdwatch dataset, using our estimated anchors makes the regression loss difference smaller compared to the traditional setup: from 0 to 0, as compared to from -0.358 to 0.223.

\subsection{ML-based Image Sampling for Training}

\begin{figure}[htbp]
    \centering
    \includegraphics[width=0.9\linewidth]{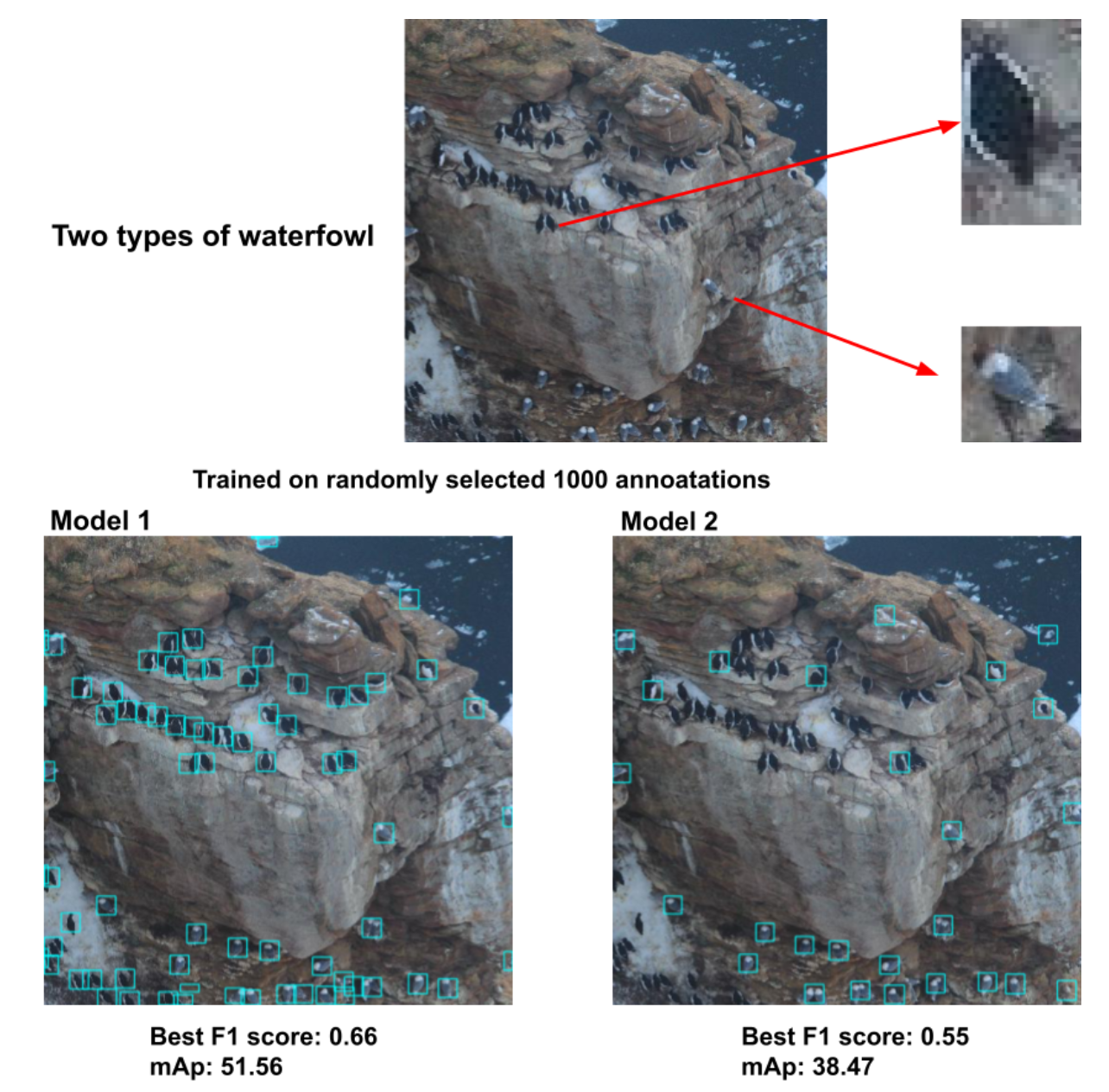}
    \caption{An example from seabirdwatch dataset showing different performance based on different training samples.}
    \label{fig:extreme_sample_example}
\end{figure}

In machine learning, it is crucial to have a balanced mix of targets in training data to train good models. For example, in the example from the seabirdwatch dataset shown in Fig.~\ref{fig:extreme_sample_example}, there are both black and white birds. When we trained a model using 1000 randomly selected  annotations, for different mix of black and white birds in the training set, the model had quite different test performances.

Inspired by the clustering methods used in data sampling\cite{cai2010unsupervised,chander2021unsupervised}, we propose a method to select a small subset of the training data that can gives us a good balanced representation of the targets in the training set. Specifically, the method is as follows. 

\begin{enumerate}
    \item Use a base model, such as ResNet-50 pre-trained on some dataset like COCO, to create features for each image. The features could be the values of the layer before the output layer. 
    \item Use a dimension reduction method, such as Principal Component Analysis (PCA) to reduce the features to smaller size.
    \item Generate additional features, such as the estimated number of bounding boxes from an existing detector, the type of image background from an existing classifier, and other useful information from the image metadata.
    \item Run a clustering method, such as K-means clustering, to group training images into clusters based on their features. Use silhouette score to find the optimal number of clusters. 
    
\item Given a  target number of images per cluster, randomly select the target number of images from each cluster. 

\end{enumerate}

\section{Experimental Results}

In this section, we present experimental results of the proposed DDR-Net, and compare its performance  with previous methods.  Two performance metrics are used:  best F1 score and  mean average precision (mAP). 
The best F1 score is the highest F1 score over different confidence thresholds.  
Mean average precision (mAP) is a standard performance metric for object detection tasks, offering a comprehensive view of model performance across all confidence thresholds and quantifying the model's ability to correctly identify and localize objects  under varying degrees of overlap and occlusion. 
In our experiments, for both F1 score and mAP calculations, an Intersection over Union (IoU) threshold of 0.3 was used and measured across all the experiments. 

In the experiments, we used public datasets used in the bird detector study \cite{weinstein2022general} for performance comparison.  The datasets are listed in Table \ref{tab:dataset_summary} here as we mentioned in previous section, GSD stands for Ground Sampling Distance and representing the image scale difference across different datasets.

\begin{table*}[h]
\centering
\caption{Summary of the public datasets of aerial images of birds used in our experiments~\cite{weinstein2022general}.}
\label{tab:dataset_summary}
\begin{tabular}{|l|l|l|l|l|}
\hline
\textbf{ Name}        & \textbf{Location}            & \textbf{No. of Training Annotations} & \textbf{No. of Test Anno.} & \textbf{GSD (cm)} \\ \hline \hline
Everglade                    & Florida, USA                 & 50870                      & 5648                       & 1.01 \\ \hline
mckellar                    & Canada                       & 1537                       & 82                         & 1.33 \\ \hline
michigan                     & Cape Cod, USA                & 40233                       & 6199                       & 0.91 \\ \hline
monash                       & Melbourne, Australia         & 9846                        & 327                        & 1.51 \\ \hline
neill                        & Utah, USA                   & 39623                       & 5119                       & 0.88 \\ \hline
newmexico                    & New Mexico, USA              & 4334                        & 283                        & 0.63 \\ \hline
palmyra                      & Palmyra Atoll, South Pacific & 1316                        & 455                        & 1.17 \\ \hline
penguins                     & Antarctic Peninsula          & 2020                        & 1505                       & 3.32 \\ \hline
pfeifer                      & South Shetland Islands, Antarctica & 43010              & 2688                       & 2.55 \\ \hline
seabirdwatch                 & North Atlantic               & 124391                      & 2362                       & 2.59 \\ \hline
\end{tabular}

\end{table*}

\subsection{Performance Comparison with Bird Detector}

In this study, we conducted a comparative analysis between our DDR-Net model and the "Bird Detector" model described in reference \cite{weinstein2022general}, a model that the original authors have openly shared. While reference\cite{weinstein2022general} outlines an evaluation strategy where the Bird Detector model is pre-trained on datas from all available training sets except for the target dataset\ref{tab:dataset_summary}, and subsequently fine-tuned with the training set from the target dataset, our experimental design adopts a different approach. For a more direct and isolated comparison, we implemented a Local-only setup, where both DDR-Net and Bird Detector were individually trained solely on the training set of each target dataset. This approach eliminates the influence of external data and allows for a clear-cut performance assessment of each model based exclusively on the target dataset's test set.

For our DDR-Net training, we started FPN part of the model with pre-trained FPN weights from COCO dataset and random weights for the rest of the model. We trained the model for 80 epochs with learning rate 0.001 and batch size 24. We used basic image augmentations, e.g, flipping, scaling, rotating, etc. The same settings were used in all experiments. 
The experiments were done on a Dell desktop computer with an RTX 4090 graphics card.  

 Table \ref{table:Bird Detector and DDR Net Performance Metrics} shows performance comparison of the two models on the datasets. Out of the nine datasets, DDR-Net is significantly better on six datasets, about the same on two datasets, and slightly worse on one. DDR-Net is from 12\% to 27.8\% faster in inference. This inference speed gain mainly resulted from fewer proposals generated by our adaptive feature map calculation as compared with standard feature map proposals.

\begin{table*}[h]
    \centering
    \caption{Performance comparison of local-only RetinaNet (i.e., Bird Detector) and DDR-Net. }
    \label{table:Bird Detector and DDR Net Performance Metrics}
    \begin{tabular}{|l|c|c|c|c|c|c|}
\hline
\textbf{Dataset} & \multicolumn{3}{c|}{\textbf{Bird Detector \cite{weinstein2022general}}} & \multicolumn{3}{c|}{\textbf{DDR-Net}} \\ \hline
 & \textbf{F1} & \textbf{mAp} & \textbf{Avg. Inference Time (s)} & \textbf{F1} & \textbf{mAp} & \textbf{Avg. Inference Time (s)} \\ \hline
mckellar & 0.57 & 45.5 & 0.343 & \textbf{0.67 (+17.5\%)} & \textbf{63.76 (+40.0\%)} & \textbf{0.302 (-12.0\%)} \\ \hline
michigan & 0.66 & 55.43 & 0.341 & \textbf{0.82 (+24.2\%)} & \textbf{82.1 (+48.0\%)} & \textbf{0.281 (-17.6\%)} \\ \hline
monash & 0.74 & 66.92 & 0.343 & \textbf{0.81 (+9.5\%)} & \textbf{78.09 (+16.7\%)} & \textbf{0.285 (-16.9\%)} \\ \hline
neill & 0.91 & 89.49 & 0.356 & \textbf{0.92 (+1.1\%)} & \textbf{90.3 (+0.9\%)} & \textbf{0.275 (-22.8\%)} \\ \hline
newmexico & 0.89 & 90.42 & 0.361 & \textbf{0.94 (+5.6\%)} & \textbf{93.19 (+3.1\%)} & \textbf{0.295 (-18.3\%)} \\ \hline
palmyra & 0.79 & 77.21 & 0.340 & {0.79 (0.0\%)} & \textbf{78.66 (+1.9\%)} & \textbf{0.286 (-15.9\%)} \\ \hline
penguins & 0.89 & 88.02 & 0.336 & 0.87(-2.2\%) & 87.03 (-1.1\%) & \textbf{0.281 (-16.4\%)} \\ \hline
pfeifer & 0.86 & 81.02 & 0.341 & \textbf{0.88 (+2.3\%)} & \textbf{89.25 (+10.2\%)} & \textbf{0.279 (-18.2\%)} \\ \hline
seabirdwatch & 0.72 & 57.49 & 0.356 & \textbf{0.83 (+15.3\%)} & \textbf{75.27 (+31.0\%)} & \textbf{0.257 (-27.8\%)} \\ \hline
\end{tabular}
\end{table*}

\subsection{Performance of ML-based Image Sampling Method}
In this experiment, we evaluate the ML-based Image Sampling method by comparing the performances of DDR-Net models trained using training sets generated by this method with those of DDR-Net models trained using unbalanced (or biased) training sets. We conducted 10 random trials, and in each trial, we restricted the training data to 1000 annotations.

To generate unbalanced training sets, we only randomly selected training examples from a subset of the clusters outputted by the k-means clustering method, instead of selecting them from all clusters.  

\begin{figure}[!t]
    \centering
    \subfloat[Results using the seabirdwatch dataset.]{
        \includegraphics[width=\linewidth]{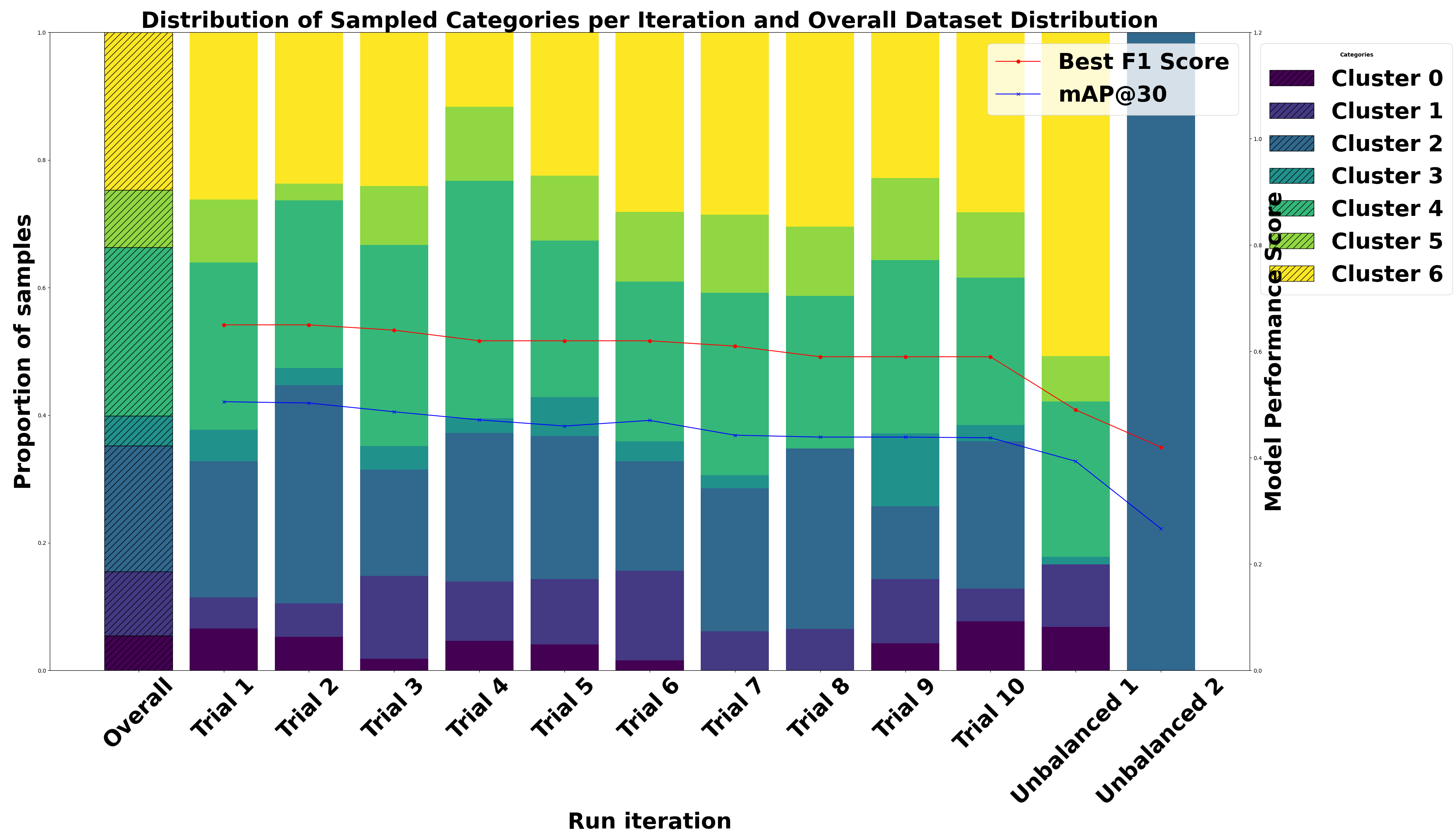}
        \label{fig:seabirdwatch}
    }
    \vspace{1em}
    \subfloat[Results using the neill dataset.]{
        \includegraphics[width=\linewidth]{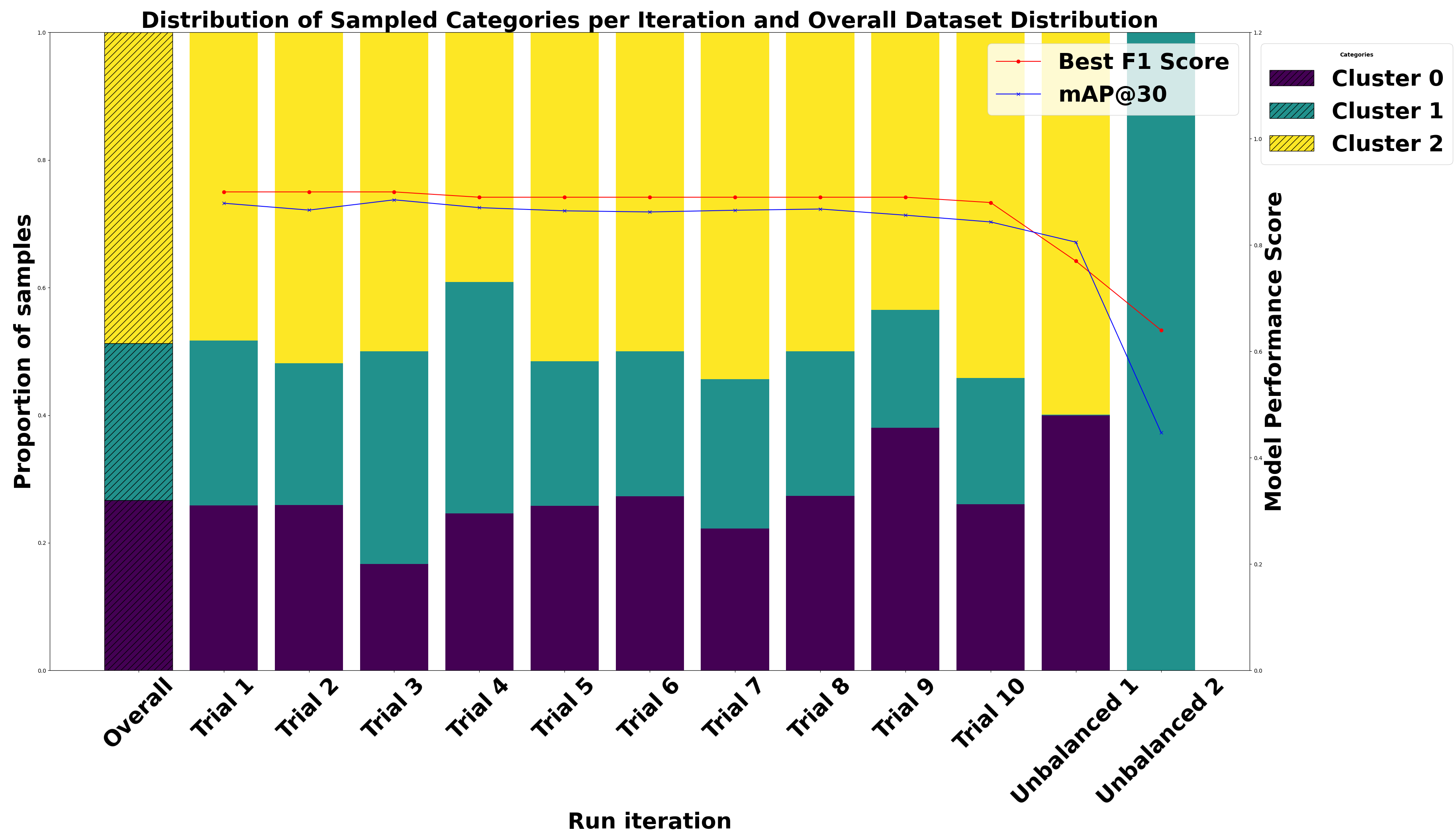}
        \label{fig:neill}
    }
    \caption{Distributions of training examples across clusters in the training sets generated by the proposed ML-based Image Sampling method and in the unbalanced training sets, and the corresponding performances (best F1 score and mAP values) of DDR-Net models trained using different training sets.}
    \label{fig:sampling_experiment}
\end{figure}

Fig. \ref{fig:sampling_experiment} shows the results on two datasets, seabirdwatch and neill,  
In each figure, the first bar shows the distribution of clusters in the training set. The number of clusters was determined by adaptive k-means clustering method. There were 9 clusters for the seabirdwatch dataset and  3 clusters for the neill dataset. 

The next 10 bars shows training example distributions generated by 
the ML-based Image Sampling method in 10 trials. The variations are due to the randomness in our sampling method. The last two bars shows the training example distributions of two unbalanced training sets.

The best F1 score and mAP values of DDR-Net model trained using each training set are shown as the red and blue curves, respectively. The results show that DDR-Net models trained using the ML-based Image Sampling method achieved consistent performances, much better than DDR-Net models trained using unbalanced training sets.

\subsection{Performance of DDR-Net Fine-Tuned with Limited
Data}
In this section, we compare the performances of DDR-Net and Bird-Detector when fine-tuned with a limited dataset. We limited the training set to a total of 1,000 annotations. First, we trained a DDR-Net model and a Bird-Detector model using the Everglades dataset. These two models were fine-tuned on the other nine datasets. 

When fine-tuning DDR-Net models, we fixed  feature map sizes and number of anchor boxes to avoid any weight mismatch between the pre-trained and fine-tuned model weights. 
During fine-tuning, we set the training epochs to 20 with learning rate 0.002, the same as \cite{weinstein2022general}. We ensured a fair comparison by keeping the sampled training data identical in each trial for both models. This experiment was repeated 10 times to get the average performance and variations. After fine-tuning, the models were evaluated on the test set of each dataset. 

\begin{figure}[h]
    \centering
    \includegraphics[width=0.9\linewidth]{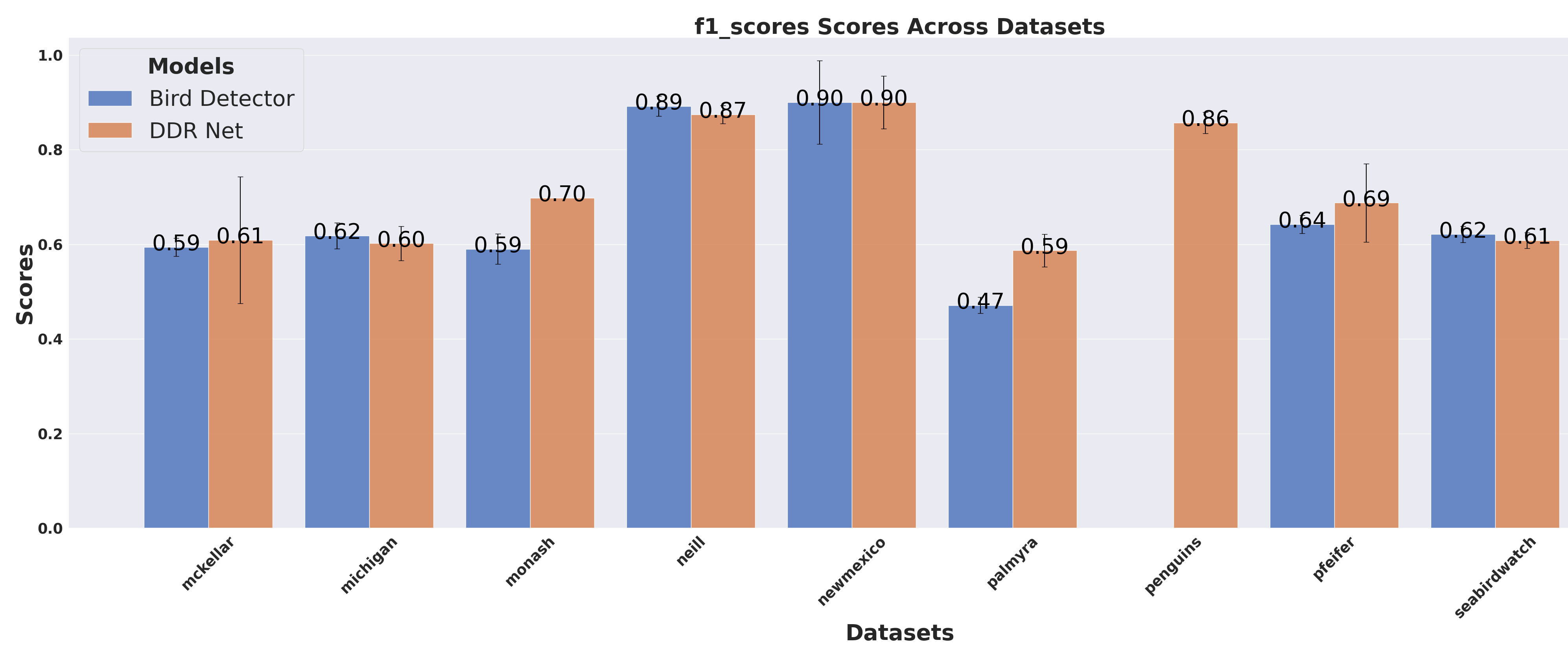}
    \caption{Performance comparison (best F1 score)  of fine-tuned Bird Detector and DDR-Net models on nine datasets.}
    \label{fig:f1_score}
\end{figure}

\begin{figure}[h]
    \centering
    \includegraphics[width=0.9\linewidth]{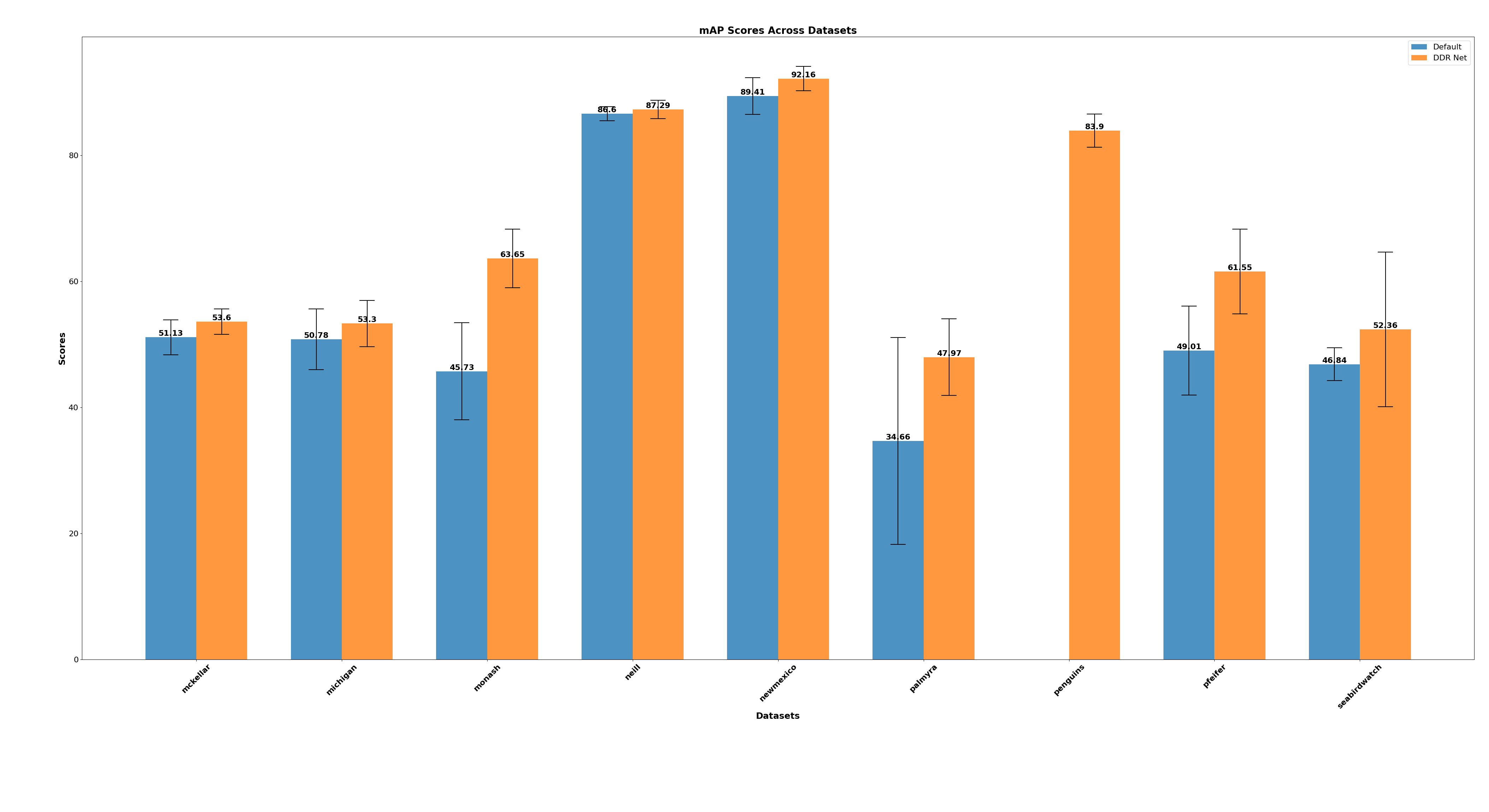}
    \caption{Performance comparison (mAp)  of fine-tuned Bird Detector and DDR-Net models on nine datasets.}
    \label{fig:mAp}
\end{figure}

Fig. \ref{fig:f1_score} and Fig. \ref{fig:mAp} compare the performance of fine-tuned Bird Detector and DDR-Net models on nine datasets. The average and standard deviation of the best F1 score and mAP for different datasets are shown. Overall, DDR-Net is significantly better than Bird Detector on 4 datasets and is similar on 5 datasets. For the penguins dataset, the Bird Detector model failed to be fine-tuned. In terms of mAP values, DDR-Net is better on all datasets. For example, on the monash dataset, DDR-Net achieved 63.65 while Bird Detector was 45.23.  

Table \ref{tab:p_values_map_percentage_gain} shows the average gains of DDR-Net models and P-values of paired-t test results between the mAP values of fine-tuned Bird Detector and DDR-Net model across the datasets, where the training data were identical between the two models. The improvement of DDR-Net is statistically significant. 



\begin{table}[h]
\centering
\caption{Average gains of DDR-Net models and P-values of paired-t test results between the mAP values of fine-tuned Bird Detector and DDR-Net model on the test sets of different  datasets.}
\label{tab:p_values_map_percentage_gain}
\begin{tabular}{|l|c|c|}
\hline
\textbf{Dataset} & \textbf{Average mAP Gain (\%)} & \textbf{P-value} \\ \hline
\hline
mckellar & $4.83$ & $1.022\times10^{-2}$ \\ \hline
michigan & $4.97$ & $8.075\times10^{-2}$ \\ \hline
monash & $39.19$ & $3.306\times10^{-2}$ \\ \hline
neill & $0.8$ & $1.451\times10^{-1}$ \\ \hline
newmexico & $3.07$ & $5.616\times10^{-2}$ \\ \hline
palmyra & $38.41$ & $5.873\times10^{-2}$ \\ \hline
pfeifer & $25.58$ & $1.474\times10^{-3}$ \\ \hline
seabirdwatch & $11.8$ & $7.893\times10^{-3}$ \\ \hline
\end{tabular}

\end{table}

\section{Conclusions}

This paper presented DDR-Net, a new data-driven deep learning architecture based on RetinaNet, along with a ML-based training data sampling method. Experimental results on a large number of datasets of aerial images on birds showed significant improvement over the state-of-the-art Bird Detector model. Fine-tuning experiments showed that DDR-Net outperformed Bird Detector using limited amount of training data. The ML-based training data sampling method was showed to be effective, which is a useful technique when dealing with large aerial imagery datasets. DDR-Net also ran faster in inference. 

In the future work, we plan to explore how DDR-Net and our sampling method function in other domains and a broader range of object detection tasks. We will improve DDR-Net and develop  smart sampling and active learning techniques  for object detection with limited data.

\section*{Acknowledgment}
ChatGPT was used to revise the writing to improve the overall readability.

\bibliographystyle{IEEEtran}
\bibliography{reference}

\begin{thebibliography}{10}
\providecommand{\url}[1]{#1}
\csname url@samestyle\endcsname
\providecommand{\newblock}{\relax}
\providecommand{\bibinfo}[2]{#2}
\providecommand{\BIBentrySTDinterwordspacing}{\spaceskip=0pt\relax}
\providecommand{\BIBentryALTinterwordstretchfactor}{4}
\providecommand{\BIBentryALTinterwordspacing}{\spaceskip=\fontdimen2\font plus
\BIBentryALTinterwordstretchfactor\fontdimen3\font minus \fontdimen4\font\relax}
\providecommand{\BIBforeignlanguage}[2]{{%
\expandafter\ifx\csname l@#1\endcsname\relax
\typeout{** WARNING: IEEEtran.bst: No hyphenation pattern has been}%
\typeout{** loaded for the language `#1'. Using the pattern for}%
\typeout{** the default language instead.}%
\else
\language=\csname l@#1\endcsname
\fi
#2}}
\providecommand{\BIBdecl}{\relax}
\BIBdecl

\bibitem{montserrat2017training}
D.~M. Montserrat, Q.~Lin, J.~Allebach, and E.~J. Delp, ``Training object detection and recognition cnn models using data augmentation,'' \emph{Electronic Imaging}, vol. 2017, no.~10, pp. 27--36, 2017.

\bibitem{kisantal2019augmentation}
M.~Kisantal, Z.~Wojna, J.~Murawski, J.~Naruniec, and K.~Cho, ``Augmentation for small object detection,'' \emph{arXiv preprint arXiv:1902.07296}, 2019.

\bibitem{zoph2020learning}
B.~Zoph, E.~D. Cubuk, G.~Ghiasi, T.-Y. Lin, J.~Shlens, and Q.~V. Le, ``Learning data augmentation strategies for object detection,'' in \emph{Computer Vision--ECCV 2020: 16th European Conference, Glasgow, UK, August 23--28, 2020, Proceedings, Part XXVII 16}.\hskip 1em plus 0.5em minus 0.4em\relax Springer, 2020, pp. 566--583.

\bibitem{lin2017focal}
T.-Y. Lin, P.~Goyal, R.~Girshick, K.~He, and P.~Doll{\'a}r, ``Focal loss for dense object detection,'' in \emph{Proceedings of the IEEE international conference on computer vision}, 2017, pp. 2980--2988.

\bibitem{zhang2022deep}
Y.~Zhang, Y.~Wang, Z.~Tang, Z.~Zhai, Y.~Shang, and R.~Viegut, ``Deep learning methods for tree detection and classification,'' in \emph{2022 IEEE 4th International Conference on Cognitive Machine Intelligence (CogMI)}.\hskip 1em plus 0.5em minus 0.4em\relax IEEE, 2022, pp. 148--155.

\bibitem{sivanandam2022tree}
P.~Sivanandam and A.~Lucieer, ``Tree detection and species classification in a mixed species forest using unoccupied aircraft system (uas) rgb and multispectral imagery,'' \emph{Remote Sensing}, vol.~14, no.~19, p. 4963, 2022.

\bibitem{ferreira2020individual}
M.~P. Ferreira, D.~R.~A. de~Almeida, D.~de~Almeida~Papa, J.~B.~S. Minervino, H.~F.~P. Veras, A.~Formighieri, C.~A.~N. Santos, M.~A.~D. Ferreira, E.~O. Figueiredo, and E.~J.~L. Ferreira, ``Individual tree detection and species classification of amazonian palms using uav images and deep learning,'' \emph{Forest Ecology and Management}, vol. 475, p. 118397, 2020.

\bibitem{tang2023new}
Z.~Tang, Y.~Liu, and Y.~Shang, ``A new gnn-based object detection method for multiple small objects in aerial images,'' in \emph{2023 IEEE/ACIS 23rd International Conference on Computer and Information Science (ICIS)}.\hskip 1em plus 0.5em minus 0.4em\relax IEEE, 2023, pp. 14--19.

\bibitem{tang2021suas}
Z.~Tang, Y.~Zhang, Y.~Wang, Y.~Shang, R.~Viegut, E.~Webb, A.~Raedeke, and J.~Sartwell, ``suas and machine learning integration in waterfowl population surveys,'' in \emph{2021 IEEE 33rd International Conference on Tools with Artificial Intelligence (ICTAI)}.\hskip 1em plus 0.5em minus 0.4em\relax IEEE, 2021, pp. 517--521.

\bibitem{pagano2009estimating}
A.~M. Pagano and T.~W. Arnold, ``Estimating detection probabilities of waterfowl broods from ground-based surveys,'' \emph{The Journal of Wildlife Management}, vol.~73, no.~5, pp. 686--694, 2009.

\bibitem{kaur2021data}
P.~Kaur, B.~S. Khehra, and E.~B.~S. Mavi, ``Data augmentation for object detection: A review,'' in \emph{2021 IEEE International Midwest Symposium on Circuits and Systems (MWSCAS)}.\hskip 1em plus 0.5em minus 0.4em\relax IEEE, 2021, pp. 537--543.

\bibitem{ren2020real}
Z.~Ren, E.~Y. Lam, and J.~Zhao, ``Real-time target detection in visual sensing environments using deep transfer learning and improved anchor box generation,'' \emph{IEEE Access}, vol.~8, pp. 193\,512--193\,522, 2020.

\bibitem{wang2022improved}
H.~Wang, B.~Wu, Y.~Wu, S.~Zhang, S.~Mei, and Y.~Liu, ``An improved yolo-v3algorithm for ship detection in sar image based on k-means++ with focal loss,'' in \emph{2022 3rd China International SAR Symposium (CISS)}.\hskip 1em plus 0.5em minus 0.4em\relax IEEE, 2022, pp. 1--5.

\bibitem{mehtab2021flexinet}
S.~Mehtab, F.~Sarwar, and W.~Yan, ``Flexinet: Fast and accurate vehicle detection for autonomous vehicles,'' in \emph{Proceedings of the 4th International Conference on Control and Computer Vision}, 2021, pp. 43--49.

\bibitem{xu2021adaptive}
Y.~Xu and Q.~Yu, ``Adaptive weighted multi-level fusion of multi-scale features: A new approach to pedestrian detection,'' \emph{Future Internet}, vol.~13, no.~2, p.~38, 2021.

\bibitem{yang2003feature}
J.~Yang, J.-y. Yang, D.~Zhang, and J.-f. Lu, ``Feature fusion: parallel strategy vs. serial strategy,'' \emph{Pattern recognition}, vol.~36, no.~6, pp. 1369--1381, 2003.

\bibitem{weinstein2022general}
B.~G. Weinstein, L.~Garner, V.~R. Saccomanno, A.~Steinkraus, A.~Ortega, K.~Brush, G.~Yenni, A.~E. McKellar, R.~Converse, C.~D. Lippitt \emph{et~al.}, ``A general deep learning model for bird detection in high-resolution airborne imagery,'' \emph{Ecological Applications}, vol.~32, no.~8, p. e2694, 2022.

\bibitem{cai2010unsupervised}
D.~Cai, C.~Zhang, and X.~He, ``Unsupervised feature selection for multi-cluster data,'' in \emph{Proceedings of the 16th ACM SIGKDD international conference on Knowledge discovery and data mining}, 2010, pp. 333--342.

\bibitem{chander2021unsupervised}
S.~Chander and P.~Vijaya, ``Unsupervised learning methods for data clustering,'' in \emph{Artificial Intelligence in Data Mining}.\hskip 1em plus 0.5em minus 0.4em\relax Elsevier, 2021, pp. 41--64.

\end{thebibliography}

\end{document}